\documentclass[journal]{IEEEtran}
\usepackage{graphicx}
\usepackage{amsmath}
\usepackage{amssymb}
\usepackage{bm}
\usepackage{algorithm}
\usepackage{algorithmic}
\usepackage{hyperref}
\hypersetup{
   colorlinks=true,
   linkcolor=blue,
   filecolor=blue,      
   urlcolor=blue,
   citecolor=blue
   }
\usepackage[dvipsnames]{xcolor}
\usepackage{tabularx}
\usepackage{multirow}
\usepackage{booktabs}

\usepackage{microtype}

\DeclareMathOperator*{\argmin}{arg\,min}
\DeclareMathOperator*{\argmax}{arg\,max}

\long\def\comment#1{}
\long\def\red#1{\bgroup\color{red}#1\egroup}

\newcommand{\xmath}[1] {\ensuremath{#1}\xspace}

\newcommand{\pbox}[1] {%
\makebox[0pt][r]{\raisebox{7mm}[0pt][0pt]{\small #1}}\ignorespaces}
\newcommand{\be} {\begin{equation}}
\newcommand{\ee}[1] {\label{#1}\end{equation}\pbox{#1}}
\newcommand{\eref}[1] {(\ref{#1})}
\newcommand{\fref}[1] {Fig.~\ref{#1}}

\newcommand{\bs}{\begin{equation}\begin{split}}

\newcommand{\st}{\hspace{2mm} \text{s.t.} \hspace{2mm}}

\newcommand\highlightReference[1]{%
  \expandafter\newcommand\csname highlightReference-#1\endcsname{}%
}
\let\oldbibitem\bibitem
\def\bibitem#1 #2\par{%
  \expandafter\ifx\csname highlightReference-#1\endcsname\relax
    \oldbibitem{#1}#2\par
  \else
    \oldbibitem{#1}\highlight{#2}\par
  \fi
}

\newcommand\highlight[1]{\textcolor{blue}{#1}}

\renewcommand{\xi}{\xmath{x_i}}

\usepackage{orcidlink}

\begin{document}

\title{PET-TURTLE: Deep Unsupervised Support Vector Machines for Imbalanced Data Clusters}

\author{Javier Salazar Cavazos\orcidlink{0009-0009-1218-9836}, \IEEEmembership{Graduate Student Member, IEEE}
\thanks{This work was supported by U.S. National Institutes of Health (NIH) under Grant R21 AG082204 and by the University of Michigan under the Rackham Merit Fellowship.}
\thanks{Javier Salazar Cavazos is with 
the Electrical and Computer Engineering (ECE) Department, University of Michigan, Ann Arbor, MI 
48109 USA.}}

\maketitle

\begin{abstract}
Foundation vision, audio, and language models enable zero-shot performance on downstream tasks via their latent representations. Recently, unsupervised learning of data group structure with deep learning methods has gained popularity. TURTLE, a state of the art deep clustering algorithm, uncovers data labeling without supervision by alternating label and hyperplane updates, maximizing the hyperplane margin, in a similar fashion to support vector machines (SVMs). However, TURTLE assumes clusters are balanced; when data is imbalanced, it yields non-ideal hyperplanes that cause higher clustering error. We propose PET-TURTLE, which generalizes the cost function to handle imbalanced data distributions by a power law prior. Additionally, by introducing sparse logits in the labeling process, PET-TURTLE optimizes a simpler search space that in turn improves accuracy for balanced datasets. Experiments on synthetic and real data show that PET-TURTLE improves accuracy for imbalanced sources, prevents over-prediction of minority clusters, and enhances overall clustering.
\end{abstract}

\begin{IEEEkeywords}
Clustering, imbalanced data, foundation models, support vector machines (SVMs), unsupervised learning. 
\end{IEEEkeywords}

\IEEEpeerreviewmaketitle

\section{Introduction}
Transfer learning takes advantage of pretrained neural networks to improve model performance in downstream tasks usually with limited data \cite{transferlearning}. Recent works show that fine-tuning an entire model has only marginal improvements compared to using a frozen backbone with a linear classifier \cite{linearclassifier1} \cite{dino} \cite{mae}. These self-supervised foundational models, trained in the CLIP or contrastive learning paradigm \cite{clip} \cite{contrastivelearning}, offer competitive performance on a variety of downstream tasks by learning general representations and using them in zero-shot settings. However, in some scenarios, data labels are unknown which makes it impossible to train a linear classifier on top of these models. Thus, one would naturally think about applying clustering methods such as K-Means \cite{kmeans} or subspace variations \cite{alpcahus} on the latent representations. It has been shown that this strategy leads to worse performance compared to weakly supervised approaches \cite{ibot}. 
\begin{figure}[t]
    \centering
    \includegraphics[width=0.80\linewidth]{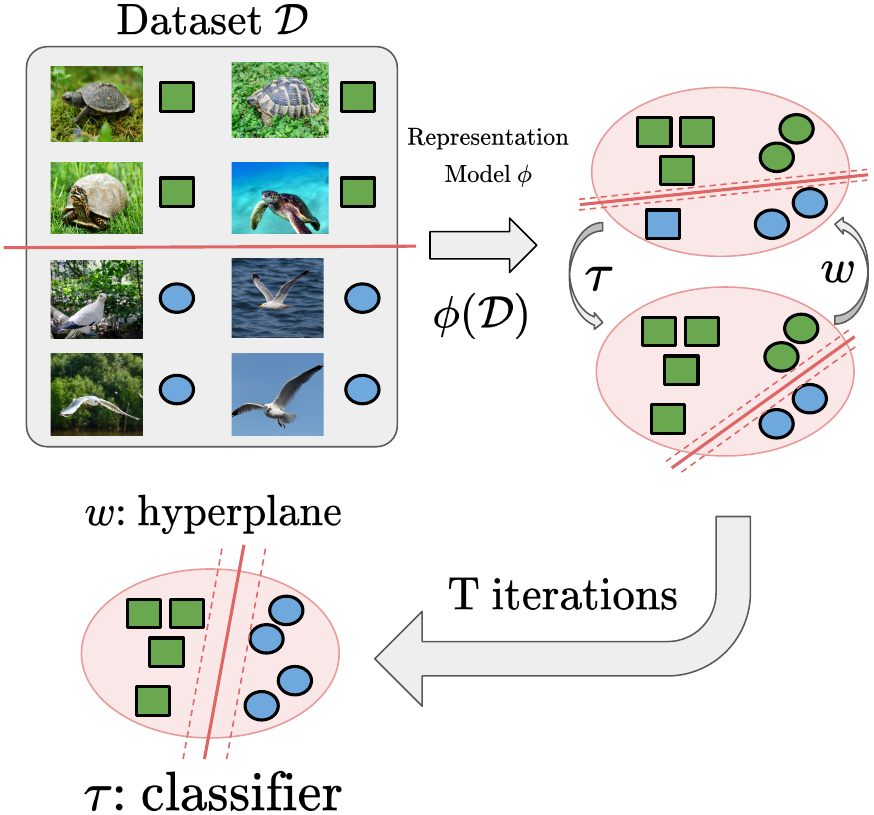}
    \caption{A visual illustration of the key idea behind unsupervised support vector machines that alternates updates between labels and hyperplane estimation.}
    \label{fig:illustration}
\end{figure}
In recent work, the TURTLE algorithm \cite{turtle} was proposed to overcome such challenges by searching for the labeling that maximizes the margin of hyperplanes to uncover data groups in an unsupervised way. Compared to other deep clustering methods such as DEC \cite{dec}, DAC \cite{dac}, DeepCluster \cite{deepcluster}, and SPICE \cite{spice}, TURTLE does not use task-specific representation learning that is typically very expensive for modern foundation models. Compared to these methods, TURTLE set the new state of the art performance in unsupervised learning for clustering. 

Diving further into the problem statement, let $\phi$ represent some backbone model such that $\phi(x) = z \in \mathbb{R}^{d}$ represents latent features belonging to $d$-dimensional space and associated with original data sample $x$ from dataset $\mathcal{D}$. Let $C$ correspond to the number of clusters. Let $\tau_{\theta}(z) : \{\mathcal{Z} = \phi(\mathcal{D})\} \rightarrow \{ 1, \ldots, C\}$ denote the classifier with continuous parameters $\theta$ that uncover the underlying labeling of $z \in \mathcal{Z}$. Specifically, let $\tau_{\theta}(z) = A_{\theta} z + b_{\theta}$ where $A_{\theta} \in \mathbb{R}^{C \times d}$ and $b_{\theta} \in \mathbb{R}^{C}$. Let $w_{\theta} \in \mathbb{R}^{C \times d}$ correspond to the hyperplane in the latent space that also includes a bias term. Then, the TURTLE optimization objective, given a single representation space, solves
\begin{align}
   \mathcal{L}_{\text{TURTLE}}(\theta) = &\sum_{z \in \mathcal{\phi(D)}} \mathcal{L}_{\text{CE}} (w_{\theta}^M z ; \sigma( \tau_{\theta} (z))) \nonumber \\
   &\st w_{\theta}^M = \Xi^{(M)} (w_{\theta}^M, \phi(\mathcal{D})) \label{eq:turtle}
\end{align}
where $\sigma(\cdot)$ denotes the softmax operation and $\mathcal{L}_{\text{CE}}(\cdot)$ is the cross entropy loss \cite{crossentropy}. The inner term $\Xi^{(M)} (w_{\theta}^M, \phi(\mathcal{D}))$ denotes an iterative optimization algorithm $\Xi$, such as gradient descent, run for $M$ steps starting from a randomly initialized $w_{\theta}^0$. In a binary classification setting with linearly separable data, \eref{eq:turtle} corresponds to unregularized logistic regression. In Ref. \cite{implicitbias}, the authors show that gradient descent when applied to \eref{eq:turtle} with the known labels induces iterates biased towards the direction of the maximum hard-margin hyperplane. Later work showed that similar results hold true in the non-separable setting for a soft-margin hyperplane \cite{implicitbias_nonseperable}. TURTLE takes advantage of the bi-level optimization problem by alternating between finding a hyperplane and using it to update labels in a similar way to K-Means, but in a support vector machine context. 

However, regularization is needed for the classifier $\tau$ to prevent the global and trivial solution where the encoder classifies all samples to the same cluster and thus technically achieves a minimum value for \eref{eq:turtle}. In this $C=1$ context, one can easily find maximum margin since the hyperplane can be far away from the single cluster data corresponding to a low cost function value as shown in Ref. \cite{hume}. Let 
\begin{equation}
    \bar{\tau}_{\theta} = \frac{1}{|\mathcal{D}|} \sum_{z \in \phi(\mathcal{D})} \tau_{\theta}(z)
\end{equation} 
be the empirical label distribution of $\mathcal{D}$ predicted by the classifier. Then, the final objective function of the TURTLE formulation is 
\begin{equation}
    \min_{\theta} \mathcal{L}_{\text{TURTLE}}(\theta) - \gamma \mathbb{H}(\bar{\tau_{\theta}}) \label{eq:turtle_final}
\end{equation}
where $\mathbb{H}(\cdot)$ corresponds to the entropy function of discrete distributions. Let $h_i \in \bar{\tau}_{\theta}$ correspond to the $i$th element of the vector $\bar{\tau}_{\theta}$. Then, $\mathbb{H}(\cdot)$ operates elementwise by 
\begin{equation}
    \mathbb{H}(h_i) =
    \begin{cases} 
          - h_i \ln{h_i} & h_i > 0 \\
          0 & h_i = 0 \\
          -\infty & h_i < 0 
          \end{cases}
\end{equation}
and this prevents the degenerate solution $C=1$. However, it comes at the cost of encouraging balanced clusters since there is equal penalty for all classes. Thus, there is an explicit, balanced cluster assumption in the TURTLE formulation that leads to worse clustering quality in the imbalanced data regime. This is a phenomena observed in self-supervised learning in general \cite{uniform_prior_assumption}. Some ways others have approached this problem, to give a few examples, is by ensemble methods that fuse information from different experts \cite{tingjin1} or by randomly encoding subtasks to find the adaptive weighting during training \cite{tingjin2}. 

In this work, we address the balanced cluster assumption by generalizing \eref{eq:turtle_final} to better handle distributions that come with imbalanced datasets leading to greater accuracy and cluster cohesion. Our new method, named PET-TURTLE, achieves higher performance in the imbalanced regime depending on the severity of data imbalance. Additionally, our method even lowers clustering error in balanced data regimes because of the sparse logit modeling of the classifier $\tau$ that limits the number of potential classes to consider for hyperplane updates to only those with higher probabilities. We conjecture that this allows for more efficient hyperplane searching that is not held down by low probability predictions. This is similar in scope to subspace learning problems where very noisy data samples can lead to a worse subspace basis estimate in principal component analysis (PCA) \cite{alpcah}. In Sec. \ref{sec:method}, the proposed PET-TURTLE algorithm is discussed in greater detail beginning with the cluster imbalance problem.

\begin{figure}[t]
    \centering
    \includegraphics[width=0.66\linewidth]{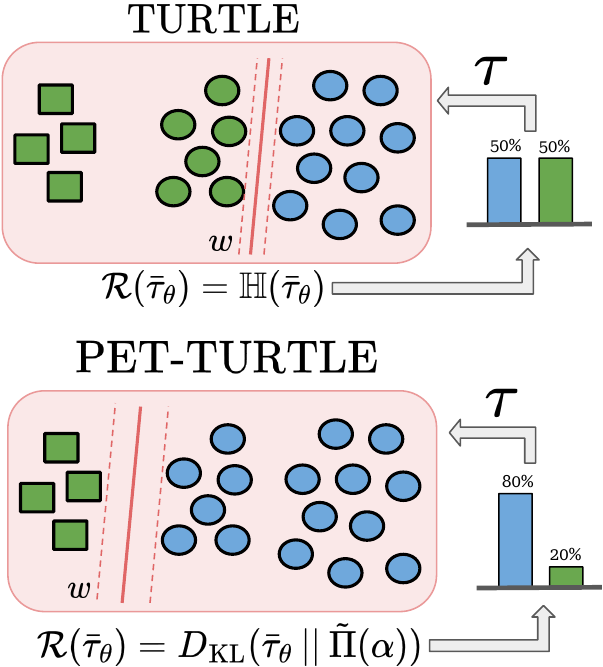}
    \caption{A visual comparison of the effects of regularization terms in the TURTLE and PET-TURTLE objective functions.}
    \label{fig:imbalance_comparison}
\end{figure}

\section{Proposed Method}
\label{sec:method}

\subsection{Prior Enforcement Term for Imbalanced Data}

To address the imbalanced cluster limitation, we consider the following class of optimization problems
\begin{equation}
    \min_{\theta} \mathcal{L}_{\text{TURTLE}}(\theta) + \gamma D_{\text{KL}}[\bar{\tau_{\theta}} \hspace{1mm} \| \hspace{1mm} \Pi ]
\end{equation}
where $D_{\text{KL}}(\cdot)$ corresponds to the KL-divergence \cite{KL} of the two distributions $\bar{\tau}_{\theta}$ and a prior data distribution $\Pi$. In the balanced cluster regime, $\Pi$ can be assumed to be uniform. In the imbalanced regime, $\Pi$ can be known for some applications, e.g., scene graph generation problems where a KL prior is added to cross entropy loss in a classification context \cite{tingjin3}. In many instances, it is more likely that the distribution is unknown, yet it can be assumed to be imbalanced. 

\begin{figure}
    \centering
    \includegraphics[width=0.85\linewidth]{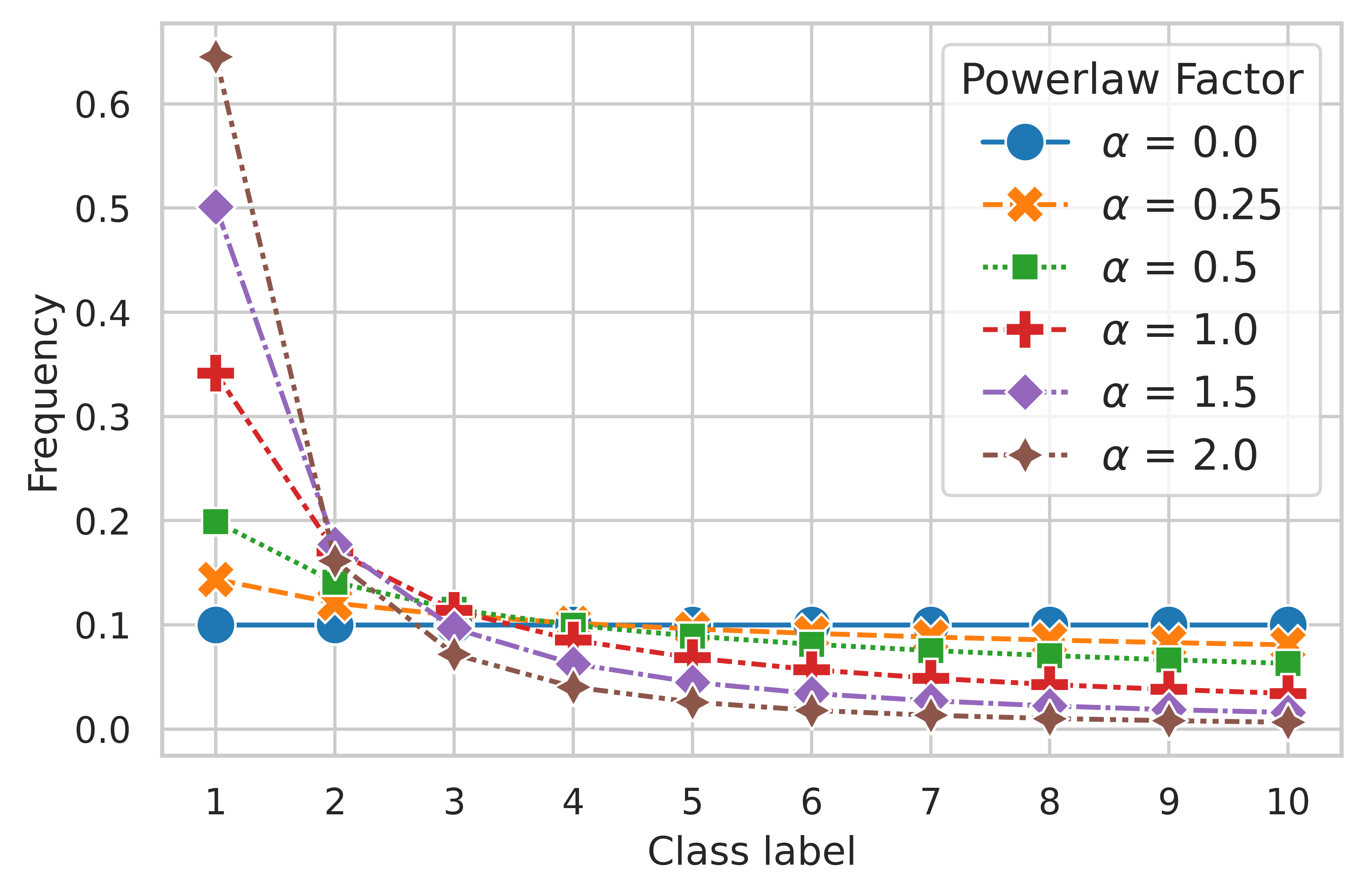}
    \caption{Probability mass functions of the power law distribution at various decay rates $\alpha$ when $C=10$ as used in Table \ref{tab:imbalance}.}
    \label{fig:histogram}
\end{figure}

We propose to use a power law distribution \cite{powerlaw} to model imbalanced data as a proxy in the unknown case. This is a natural choice for imbalanced distributions due to the wide assortment of applications in physics, biology, medicine, chemistry, astronomy, and economics \cite{powerlaw}. The power law probability mass function, given decay factor $\alpha \in \mathbb{R}_{\geq 0}$, is defined as
\begin{equation}
    p_{\text{powerlaw}}(c; \alpha) = \frac{c^{-\alpha}}{\sum_{x=1}^{C+1} c^{-\alpha}}
\end{equation}
where $c \in \{1, \ldots, C \}$. Let $\tilde{\Pi}(\alpha) \in \mathbb{R}_{\geq 0}^{C}$ represent the vectorized mass function for the power law distribution such that $\forall c, \hspace{1mm} p_{\text{powerlaw}}(c; \alpha) \in \tilde{\Pi}(\alpha)$. See \fref{fig:histogram} for a plot illustrating the power law distribution when $C = 10$ at different $\alpha$ values. 

Since the distribution has a statistical parameter $\alpha$ that describes data imbalance, one could utilize domain knowledge or known labels to pick a reasonable choice. For applications where the imbalance is not known, $\alpha$ can be estimated. One can find optimal $(\gamma,\alpha)$ values by using estimated labels $\{y_1, \ldots, y_N\}$ from $y_i = \argmax \tau_{\theta}(\phi(x_i)) \hspace{2mm} \forall i \in \mathcal{D}$ and measure the generalization error of the  linear classifier trained on those estimated labels in a cross validation setting as similarly done for $\gamma$ in Ref. \cite{turtle}. This process does not require ground truth labels and instead measures hyperplane margin given the algorithm derived labels. Because of this, no prior knowledge is necessary besides suspecting imbalanced data in a given problem setting.

\subsection{Sparse Logits for Hyperplane Estimation}

From \eref{eq:turtle}, observe that $\mathcal{L}_{\text{CE}} (w_{\theta}^M z ; \sigma(\tau_{\theta} (z)))$ is using soft labels to find the optimal hyperplane. This is because a discrete search space contains $\mathcal{O}(C^N)$ possible labelings, which is an NP-hard problem \cite{nphard}. The search space is restricted with continuous parameters $\theta$ in $\tau_{\theta}(x) =A_{\theta} x + b_{\theta}$ to enable efficient gradient optimization. However, due to the softmax operation $\sigma(\tau_{\theta}(z))$, every logit plays a role in updating the hyperplane; even low value logits that are unlikely to be the correct label for a fixed data point. This is due to the full support of the softmax function, and we conjecture from experimentation that this may cause a non-ideal estimation of the hyperplane given the results in Table \ref{tab:results}. 

Instead, we propose to filter low value logits by applying the sparsemax function \cite{sparsemax} defined as solving
\begin{equation}
\text{sparsemax}(z) = \argmin_{p \in \Delta^{C-1}} \| p - z \|_2^2
\end{equation}
where $\Delta^{C-1}$ corresponds to the probability simplex. This operation returns the Euclidean projection of $z$ onto the simplex. 
Simplex projection is a problem with efficient solutions that involve soft thresholding smaller values with automatically chosen thresholds \cite{projectsimplex}. 
We use this idea in the loss function and propose a modified clustering loss with sparse simplex projection as
\begin{align}
   \mathcal{L}_{\text{SSP}}(\theta) = &\sum_{z \in \mathcal{\phi(D)}} \mathcal{L}_{\text{CE}} (w_{\theta}^M z ; \text{sparsemax}( \tau_{\theta} (z))) \nonumber \\
   &\st w_{\theta}^M = \Xi^{(M)} (w_{\theta}^M, \phi(\mathcal{D})) \label{eq:sparse_turtle}
\end{align}
with the overall objective that includes the prior term being 
\begin{equation}
    \min_{\theta} \mathcal{L}_{\text{SSP}}(\theta) + \gamma D_{\text{KL}}[\bar{\tau_{\theta}} \hspace{1mm} \| \hspace{1mm} \tilde{\Pi}(\alpha) ] . \label{eq:pet_turtle}
\end{equation}
This variant is denoted as \textbf{PET-TURTLE} (\textbf{P}rior \textbf{E}nforcement \textbf{T}erm \textbf{TURTLE}).
For completeness, algorithm pseudocode for PET-TURTLE is provided in Alg. \ref{alg:turtle}. Additionally, Pytorch code is provided at \href{https://github.com/javiersc1/pet-turtle}{github.com/javiersc1/pet-turtle} for the method.

\begin{algorithm}
\caption{PET-TURTLE}
    \begin{algorithmic}
        \STATE \textbf{Input}: Dataset $\mathcal{D}$, representation model $\phi$, classes $C$, and regularization parameters $(\gamma, \alpha)$ \\
        \STATE \textbf{Parameters}: Iterations $T=6000$, learning rate $\eta=10^{-3}$, and inner steps $M=10$ \\
        \STATE Extract latent variables $\mathcal{Z} = \phi(\mathcal{D})$
        \FOR{$t=1$ to $T$}
            \STATE Sample mini-batch representations $z \sim \mathcal{Z}$
            \\ {\color{OliveGreen} // update plane given fixed logits $\text{sparsemax}(\tau(z))$ }
            \STATE $w^M_{\theta} \leftarrow w^M_{\theta} - \eta \frac{\partial }{\partial w_{\theta}} \left[ \text{cost function in \eref{eq:pet_turtle}} \right]$ for $1,\ldots,M$
            \\ {\color{OliveGreen} // update classifier given fixed hyperplane $\omega^M$} 
            \STATE $\tau_{\theta} \leftarrow \tau_{\theta} - \eta \frac{\partial }{\partial \tau_{\theta}} \left[ \text{cost function in \eref{eq:pet_turtle}} \right ] $ 
            \STATE \algorithmicif\ warm-start \algorithmicthen 
            \\ \hspace{2mm} {\color{OliveGreen} // use latest plane to initialize for next iteration $w^0_{\theta}$} 
            \\ \hspace{2mm} update start point $w_{\theta}^0 \leftarrow w^M_{\theta}$ \
        \ENDFOR
        \STATE \textbf{Output}: Cluster labels $\argmax \tau_{\theta}(\mathcal{Z})$
    \end{algorithmic}
\label{alg:turtle}
\end{algorithm}

\section{Experiments \& Results}

\subsection{Experimental Setup}

For all of our experiments, CLIP-RN50x64 \cite{clip} is used for feature extraction meaning only a single representation space is used. We compare PET-TURTLE against TURTLE \cite{turtle} and include the K-Means++ algorithm \cite{kmeans} to establish a baseline. Additionally, linear probing \cite{simclr}, the \emph{supervised} linear classifier method, is included to determine the highest accuracy possible assuming the labels are known for the associated latent features. It is worth noting that, in general, clustering is a harder problem than classification so results can vary greatly depending on the difficulty of the problem. For all results, the average and standard deviation of 10 trials is reported with different model seed initializations, where applicable, for transparency. Furthermore, paired t-tests were conducted between TURTLE and PET-TURTLE trials, and results with p-values less than 0.01 are indicated with the ``\textbf{*}'' symbol.

\begin{table}
\scriptsize
\centering
\resizebox{0.98\columnwidth}{!}{
\begin{tabular}{ll|cccc|c}
    & &\hspace{-1.2em} & \rotatebox[origin=lb]{90}{\smash{K-means++ \cite{kmeans}}} & \rotatebox[origin=lb]{90}{\smash{TURTLE \cite{turtle}}} & \rotatebox[origin=lb]{90}{\smash{PET-TURTLE}}  & \hspace{0.5em}\rotatebox[origin=lb]{90}{\smash{Linear Probe \cite{simclr}}} \\
    \midrule
    \multirow{5}{0em}{\rotatebox[origin=c]{90}{CIFAR10-PL}}
        & Powerlaw($\alpha=0.25$) & \hspace{-1.2em} & \hspace{-0.9em}65.5 $\pm$ 4.0\hspace{-0.4em} & \hspace{-0.9em}72.8 $\pm$ 0.3\hspace{-0.4em} & \hspace{-0.9em}\textbf{78.7*} $\pm$ 2.6\hspace{-0.4em} & \hspace{-0.5em}93.8\hspace{-0.4em} \\
        & Powerlaw($\alpha=0.50$) & \hspace{-1.2em} & \hspace{-0.9em}60.3 $\pm$ 4.3\hspace{-0.4em} & \hspace{-0.9em}68.2 $\pm$ 1.3\hspace{-0.4em} & \hspace{-0.9em}\textbf{74.0*} $\pm$ 2.0\hspace{-0.4em} & \hspace{-0.5em}93.8\hspace{-0.4em} \\
        & Powerlaw($\alpha=1.0$) & \hspace{-1.2em} & \hspace{-0.9em}50.9 $\pm$ 2.8\hspace{-0.4em} & \hspace{-0.9em}54.9 $\pm$ 3.0\hspace{-0.4em} & \hspace{-0.9em}\textbf{71.5*} $\pm$ 3.5\hspace{-0.4em} & \hspace{-0.5em}94.6\hspace{-0.4em} \\
        & Powerlaw($\alpha=1.50$) & \hspace{-1.2em} & \hspace{-0.9em}42.5 $\pm$ 3.0\hspace{-0.4em} & \hspace{-0.9em}48.3 $\pm$ 4.8\hspace{-0.4em} & \hspace{-0.9em}\textbf{67.1*} $\pm$ 4.1\hspace{-0.4em} & \hspace{-0.5em}95.0\hspace{-0.4em} \\
        & Powerlaw($\alpha=2.0$) & \hspace{-1.2em} & \hspace{-0.9em}32.8 $\pm$ 2.6\hspace{-0.4em} & \hspace{-0.9em}42.8 $\pm$ 5.0\hspace{-0.4em} & \hspace{-0.9em}\textbf{60.6*} $\pm$ 3.9\hspace{-0.4em} & \hspace{-0.5em}96.2\hspace{-0.4em} \\
    \bottomrule
    \vspace{1mm}
    \end{tabular}
    }
    \caption{Accuracy results (\%) of clustering methods on the CIFAR10-PL dataset at various power law decay rates.}
    \label{tab:imbalance}
\end{table}

\subsection{Synthetic Results}

We generated a power law imbalanced dataset based on CIFAR10 \cite{cifar10} at various decay rates $\alpha \in \{ 0.25, 0.50, 1.0, 1.50, 2.0 \}$ and denote this variant as ``CIFAR10-PL''. Since TURTLE and PET-TURTLE have a regularization parameter $\gamma$, cross validation is used to find the one that leads to the lowest generalization error in the set $\gamma \in \{1, 5, 10, 25, 50, 100, 250, 500\}$. Because the decay rate $\alpha$ is known in this synthetic experiment, it is fixed for PET-TURTLE. Table \ref{tab:imbalance} compares PET-TURTLE against the other methods at different $\alpha$ decay rates. From this result, it is clear that PET-TURTLE achieves higher clustering accuracy, especially at higher data imbalance levels.

\begin{figure}
    \centering
    \includegraphics[width=0.99\linewidth]{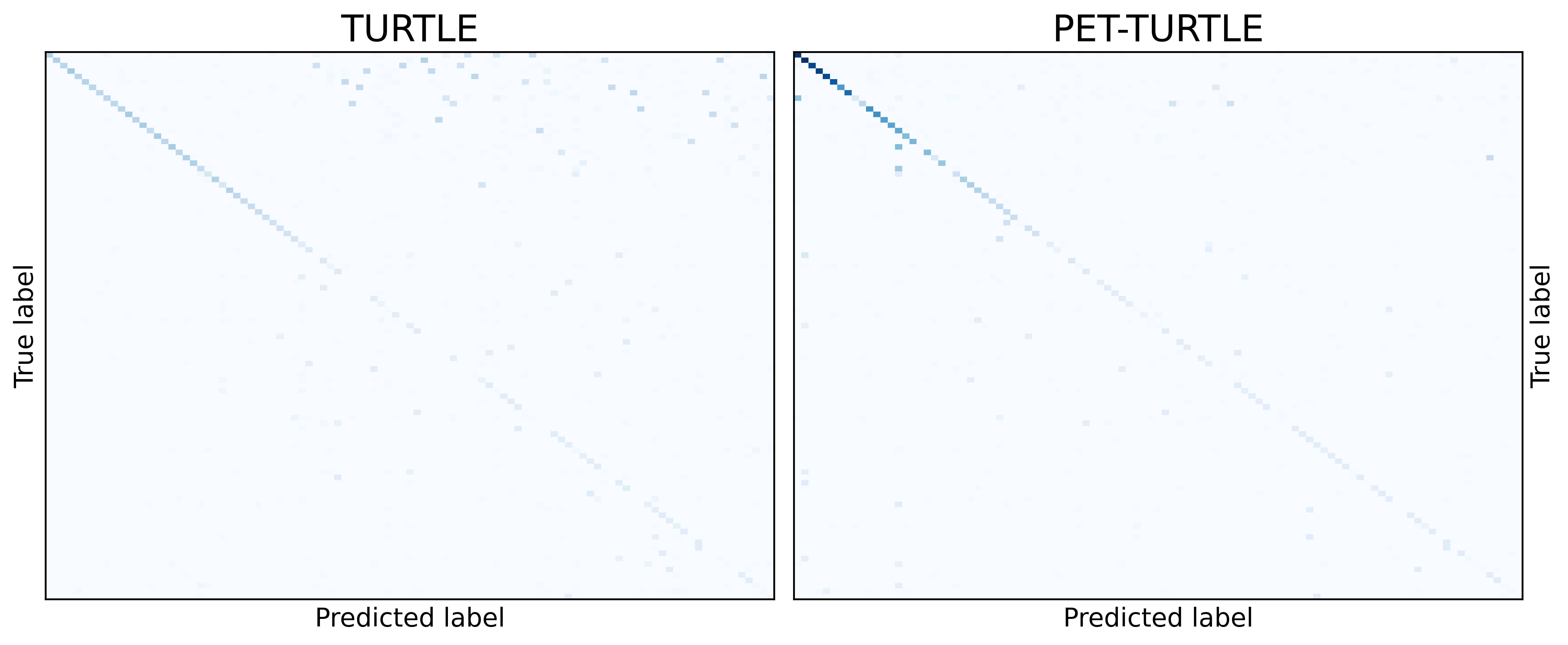}
    \caption{Confusion matrices of the TURTLE and PET-TURTLE methods on the Food101-PL dataset ($C=101$) with fixed decay rate $\alpha = 1.0$.}
    \label{fig:confusion}
\end{figure}

Similarly, a synthetic power law dataset based on Food101 \cite{food101} is generated at a fixed $\alpha = 1.0$ that contains $C=101$ classes and denote this as ``Food101-PL''. In this experiment, only TURTLE and PET-TURTLE are compared to study the effects of over-prediction. Confusion matrices between the ground truth labels and the predicted cluster labels are generated in \fref{fig:confusion}. The ``Hungarian method'' \cite{hungarian} is used to solve the cluster label permutation problem to ensure optimal correspondence between the true labels and the cluster labels. In \fref{fig:confusion}, the TURTLE confusion matrix contains many elements in the upper triangular region indicating that many samples that belong in the majority clusters were clustered into the minority clusters, thus leading to worse clustering. This is in contrast to PET-TURTLE that not only exhibits less off-diagonal elements, but also has a more correct color shift from dark blue to light blue in regards to the counts belonging to each cluster, which is expected in an imbalanced data problem.

\subsection{Real Data Results}

Lastly, real data comparisons are made on balanced and imbalanced datasets in Table \ref{tab:results}. For the balanced datasets, the following are used: Caltech101 \cite{caltech101}, CIFAR10 \cite{cifar10}, DTD \cite{dtd}, EuroSAT \cite{eurosat}, and Food101 \cite{food101}. For the imbalanced datasets, the following are used: blood cell microscope (Blood) \cite{blood}, dermatoscope (Derma) \cite{derma}, iNaturalist 2017 \cite{inaturalist}, retinal OCT (OCT) \cite{oct}, axial abdominal CT (OrganA) \cite{organ}, and kidney cortex microscope (Tissue) \cite{tissue}. The MedMNIST \cite{medmnist} package is used for data loading the medical imaging datasets. The number of classes $C$ is known for all datasets. Cross validation is done similarly as before. However, in this instance, $\alpha$ is unknown and must be estimated. Grid search is used for $\alpha \in \{0.01, 0.05, 0.10, 0.25, 0.50, 0.75, 1.0, 1.25, 1.50, 1.75, 2.0\}$ and the pair $(\gamma, \alpha)$ that achieves the lowest validation error is selected. On average, in the balanced data regime, PET-TURTLE achieves a $\sim$3\%  accuracy improvement relative to TURTLE due to the sparse logits idea used for hyperplane estimation. In the imbalanced regime, on average, PET-TURTLE achieves a $\sim$15\% accuracy improvement relative to TURTLE due to the prior enforcement term. It is important to note that the imbalanced datasets do not strictly follow a power law distribution so further improvement in clustering quality may be possible. In a way, this tests the robustness of our method against distribution mismatch. 

\begin{table}
\scriptsize
\centering
\resizebox{0.99\columnwidth}{!}{
\begin{tabular}{ll|c|ccc|c}
    & Dataset & \rotatebox{90}{$C$ (num. of classes)} & \rotatebox{90}{K-means++ \cite{kmeans}} & \rotatebox{90}{TURTLE \cite{turtle}} & \rotatebox{90}{PET-TURTLE} & \rotatebox{90}{Linear Probe \cite{simclr}} \\
    \midrule
    \multirow{5}{*}{\rotatebox[origin=c]{90}{Balanced}}
        & Caltech \cite{caltech101} & 101 & $76.6 \pm 4.5$ & $85.3 \pm 0.9$ & $\textbf{88.2*} \pm 0.5$ & $96.9$ \\
        & CIFAR \cite{cifar10} & 10 & $61.7 \pm 4.2$ & $76.3 \pm 0.1$ & $\textbf{80.9*} \pm 1.6$ & $94.1$ \\
        & DTD \cite{dtd} & 47 & $48.2 \pm 1.8$ & $57.6 \pm 0.9$ & $\textbf{59.4} \pm 1.5$ & $82.5$ \\
        & EuroSAT \cite{eurosat} & 10 & $60.0 \pm 6.4$ & $77.6 \pm 0.2$ & $\textbf{80.4*} \pm 1.6$ & $95.2$ \\
        & Food \cite{food101} & 101 & $63.1 \pm 1.6$ & $85.4 \pm 1.1$ & $\textbf{88.1*} \pm 1.0$ & $94.4$ \\
    \midrule 
    \multirow{6}{*}{\rotatebox[origin=c]{90}{Imbalanced}}
        & Blood \cite{blood} & 8 & $44.9 \pm 0.7$ & $48.1 \pm 1.8$ & $\textbf{56.1*} \pm 2.7$ & $96.1$ \\
        & Derma \cite{derma} & 7 & $25.4 \pm 0.7$ & $34.1 \pm 0.3$ & $\textbf{67.1*} \pm 0.3$ & $81.8$ \\
        & iNaturalist \cite{inaturalist} & 13 & $43.6 \pm 1.7$ & $55.9 \pm 3.7$ & $\textbf{77.5*} \pm 1.0$ & $98.1$ \\
        & OCT \cite{oct} & 4 & $46.2 \pm 0.9$ & $54.1 \pm 0.7$ & $\textbf{61.4*} \pm 0.7$ & $93.4$ \\
        & OrganA \cite{organ} & 11 & $43.5 \pm 2.8$ & $45.0 \pm 1.1$ & $\textbf{51.9*} \pm 2.4$ & $89.9$ \\
        & Tissue \cite{tissue} & 8 & $26.3 \pm 0.2$ & $30.6 \pm 1.8$ & $\textbf{38.7*} \pm 1.2$ & $60.5$ \\ 
    \bottomrule
    \vspace{0.1mm}
\end{tabular}
}
\caption{Accuracy results (\%) of clustering methods on real balanced and imbalanced image datasets.}
\label{tab:results}
\end{table}

\section{Conclusion}

In summary, this work presents a principled extension to the TURTLE clustering algorithm, addressing its limitations under imbalanced data settings by introducing a prior distribution term. This approach demonstrates robust improvements in clustering accuracy across both synthetic and real world scenarios, particularly in settings with significant class imbalance. These empirical gains highlight PET-TURTLE's practical relevance for unsupervised clustering applications atop foundation models, paving the way for more reliable clustering methods in the presence of imbalanced data distributions.

However, this proposed method relies on features extracted from a foundation model trained in some self-supervised paradigm. Thus, if one applies TURTLE or PET-TURTLE directly to ambient space data ($x \in \mathcal{D}$) that exhibits nonlinear structure, then both methods would not work well since they expect a more linearly separable space; such as one induced by foundation models. On a related note, this is why \emph{kernel} SVMs are more powerful than traditional SVMs for nonlinear settings. It would be interesting to explore an extension to PET-TURTLE in this nonlinear regime. This could be useful in a foundation model context since there is likely some nonlinearity in the latent space ($z \in \phi(\mathcal{D})$) depending on the difficulty of the problem or training distribution, that makes it more challenging to find the optimal hyperplane. Further exploration on this topic would be an interesting direction for future work.

\section{Impact Statement}

The main goal of this work is to advance the field of clustering and the proposed method relies on representation spaces of foundation models. It is known that these models inherit biases embedded in the training data \cite{bias}. Because of this, caution is recommended when using PET-TURTLE in sensitive areas such as medical imaging.

\newpage

\bibliographystyle{IEEEtran}
\bibliography{refs}

\end{document}